\documentclass[letterpaper,twocolumn,fleqn]{article} 

\usepackage{ist}
\usepackage{booktabs,multirow}
\usepackage[hidelinks]{hyperref}
\usepackage{graphicx}
\usepackage{multirow}
\usepackage{inputenc}
\usepackage[font={footnotesize}, textfont=sf]{subfig}
\usepackage{amsmath}
\usepackage{xcolor}
\usepackage{amssymb}
\usepackage[symbol]{footmisc}
 
\usepackage[hidelinks]{hyperref}
\usepackage{array}      % for better column definitions
\usepackage{multirow}   % for cells spanning multiple rows
\usepackage{tabularx}   % for tables with adjustable-width columns
\usepackage{longtable}  % for tables that can span multiple pages

\ifx00 %ifx01 to turn off and ifx00 to turn on 
\newcommand{\SKS}[1]{\textcolor{red}{[Sushil: #1]}}
\else 
\newcommand{\SKS}[1]{\textcolor{red}{}}
\fi
\title{Optimizing Ego Vehicle Trajectory Prediction: The
Graph Enhancement Approach}

\author{Sushil Sharma$^1$, Aryan Singh$^1$, Ganesh Sistu$^1$, Mark Halton$^1$, and Ciarán~Eising$^1$\\
$^1$Data-Driven Computer Engineering Group, Dept. of Electronic and Computer Engineering, University of Limerick, Ireland}
\date{} % date has an empty field.

\begin{document} 

\maketitle 

\thispagestyle{empty} % prevents the first page to be numbered

%%%%%%%%%%%%%%%%%%%%%%%%%%%%%%%%%%
% Abstract
%%%%%%%%%%%%%%%%%%%%%%%%%%%%%%%%%%

\begin{abstract}
Predicting the trajectory of an ego vehicle is a critical component of autonomous driving systems. Current state-of-the-art methods typically rely on Deep Neural Networks (DNNs) and sequential models to process front-view images for future trajectory prediction. However, these approaches often struggle with perspective issues affecting object features in the scene. To address this, we advocate for the use of Bird's Eye View (BEV) perspectives, which offer unique advantages in capturing spatial relationships and object homogeneity. In our work, we leverage Graph Neural Networks (GNNs) and positional encoding to represent objects in a BEV, achieving competitive performance compared to traditional DNN-based methods. While the BEV-based approach loses some detailed information inherent to front-view images, we balance this by enriching the BEV data by representing it as a graph where relationships between the objects in a scene are captured effectively.
% In this study, we present a comprehensive approach for ego vehicle trajectory prediction using computer vision and deep learning techniques. Leveraging a semantic segmentation model, we extract both bounding box coordinates and segmentation mask data from Bird's Eye View (BEV) images. The segmentation masks enable the precise isolation of specific image regions, which are subsequently utilized for feature extraction via a ResNet-18 network. These extracted features form the foundation for a K-nearest neighbors (KNN) algorithm, facilitating the establishment of edges connecting bounding boxes. The culmination of this process results in the construction of a graph, with bounding box features serving as nodes and edges derived from the KNN approach. To harness spatio-temporal patterns within this graph representation, we employ a Graph Neural Network (GNN) model. The GNN learns and captures the intricate spatio-temporal features critical for trajectory prediction. Finally, the acquired features are passed through multiple layers of Long Short-Term Memory (LSTM) networks to predict the trajectory of the ego vehicle, enhancing the accuracy and reliability of future path forecasting.
\end{abstract}

%%%%%%%%%%%%%%%%%%%%%%%%%%%%%%%%%%%%
% Overall Document Guidelines: Head
%%%%%%%%%%%%%%%%%%%%%%%%%%%%%%%%%%%%
%\section{Introduction}
\section{INTRODUCTION}
\vspace*{1mm}
\label{sec:intro}

The rapid advancement of autonomous driving technology has ushered in a new era of transportation, promising safer and more efficient roadways. Central to the realization of this vision is the accurate prediction of ego vehicle trajectories, a critical component for autonomous systems to navigate and interact with their environment. Accurate trajectory prediction hinges on the ability to comprehend complex spatial and temporal relationships within dynamic scenes. In this context, trajectory prediction plays an important role in enhancing the safety, efficiency, and overall performance of autonomous driving systems. It enables such vehicles to proactively anticipate the actions of various road users, including pedestrians, cyclists, and other vehicles, thereby reducing the risk of potential collisions \cite{bib7, bib8} and ensuring the navigation of possibly complex traffic scenarios. Additionally, trajectory prediction enables autonomous vehicles to optimize driving behaviour, potentially resulting in smoother lane changes \cite{bib9} and more efficient merging, ultimately contributing to the overall improvement of traffic flow and congestion reduction \cite{bib10}.
Furthermore, trajectory prediction is important for facilitating effective communication and interaction between autonomous vehicles, human drivers, and pedestrians. By adhering to predictable behaviour patterns, autonomous vehicles can establish trust among road users \cite{bib11, bib12}. This multifaceted role of trajectory prediction underscores its critical importance in autonomous driving, making it an area of continuous research and innovation.

\begin{figure}[t]
    \centering
    \includegraphics[width=0.52\textwidth]{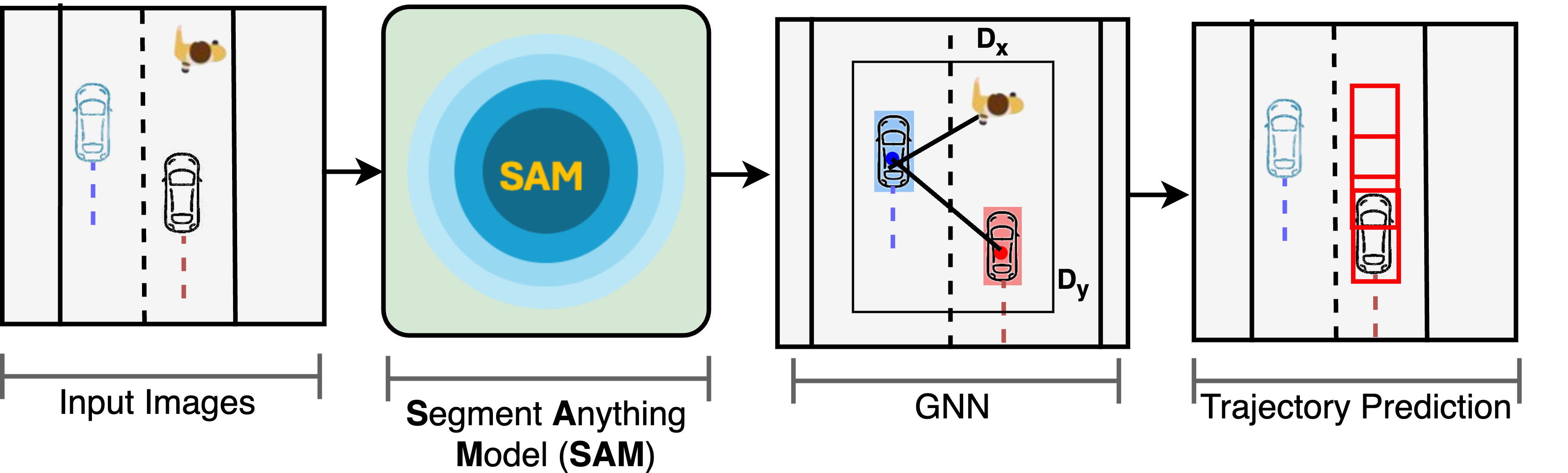}
    \vspace*{1mm}
    \caption{\textbf{Our Overview:} Segment anything model \cite{sam} extracts bounding box info. GNN processes the graph for spatial feature relations, predicting ego vehicle trajectory with LSTM layers.}
    \label{fig:image1}
\end{figure}

\begin{figure*}[t]
    \centering
   % \captionsetup{font={normalsize}, textfont=sf} 
    \includegraphics[width=0.99\textwidth]{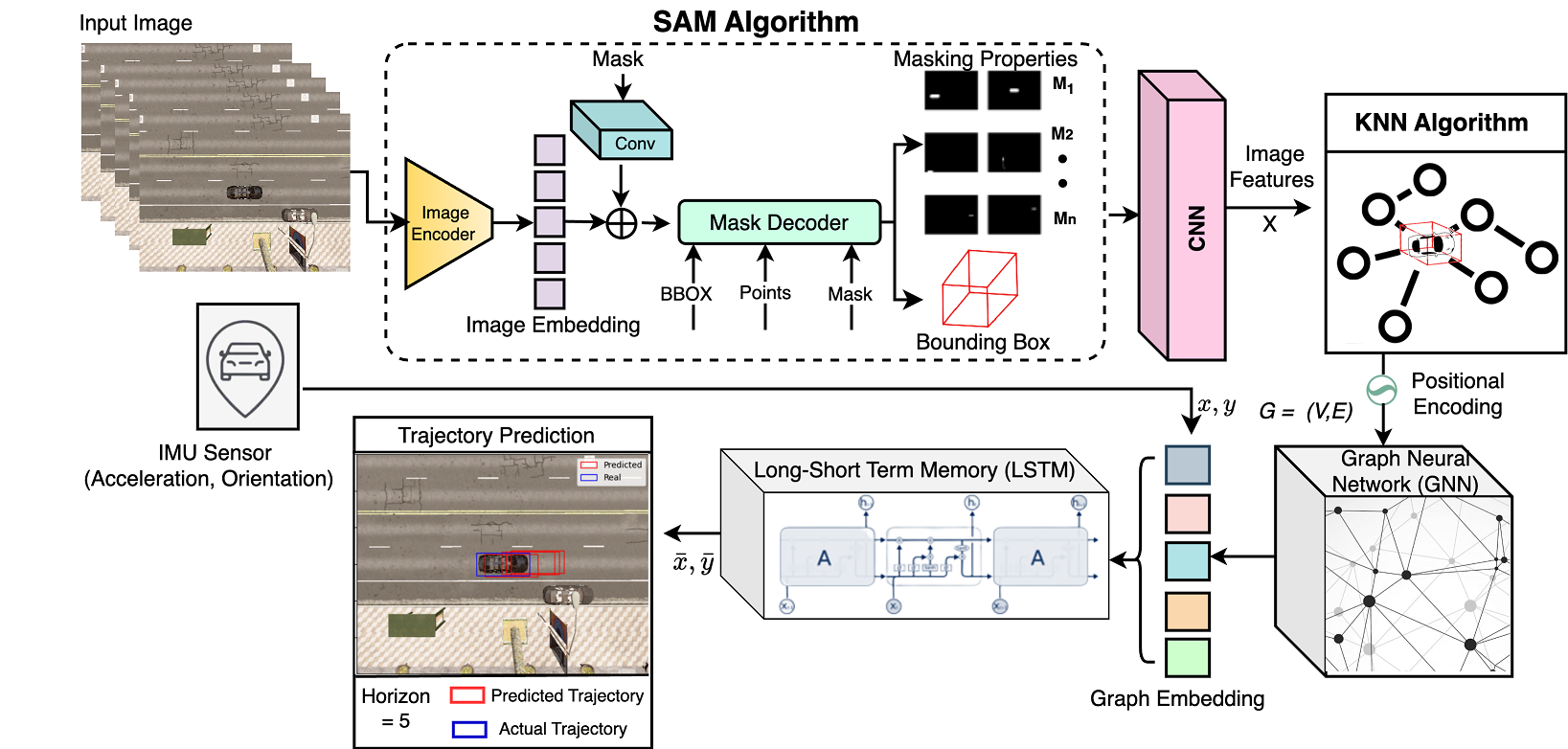}
    \vspace*{4mm}
    \caption{\textbf{Our proposed architecture:} Semantic segmentation derives bounding box coordinates and mask details from a BEV, this information is then utilized by a DNN to inform a KNN, which establishes connections between the boxes to create a graph. A GNN, enhanced with positional encoding, captures spatial features, while LSTM layers integrate temporal dynamics for the prediction of the ego vehicle's trajectory.}
    \label{fig:image2}
\end{figure*}

In this study, we present an approach for ego vehicle trajectory prediction using computer vision and deep learning techniques as shown in Figure \ref{fig:image1}. Leveraging a semantic segmentation model, we extract both bounding box coordinates and segmentation mask data from Bird's Eye View (BEV) images. The segmentation masks enable the precise isolation of specific image regions, which are subsequently utilized for feature extraction via a ResNet-18 network. These extracted features are used in a K-nearest neighbor (KNN) \cite{knn} algorithm, facilitating the establishment of edges connecting bounding boxes. Thus a graph is constructed, with bounding box features serving as nodes and edges derived from the KNN approach. To harness spatio-temporal patterns within this graph representation, we employ a Graph Neural Network (GNN) \cite{gcn} model, which learns spatio-temporal features critical for trajectory prediction. Finally, the acquired features are passed through multiple layers of Long Short-Term Memory (LSTM)\cite{lstm} networks to predict the trajectory of the ego vehicle. Our main contributions are listed below:

\begin{itemize}
    \item Separation of spatial and temporal features, addressing the \textbf{limitations of conventional CNN-based approaches} in trajectory prediction \cite{cnn_lim}.
   % \item Introducing a \textbf{novel dataset}\footnote{\textcolor{darkgray}{\href{https://drive.google.com/drive/folders/1JPb64bGV88ymZkJrUBaKQg12tToZVF7T?usp=sharing}{Dataset: https://drive.google.com/drive/folders/1JPb64bGV88ymZkJrUBaKQg1\\2tToZVF7T?usp=sharing}}} to inspire the research community to explore the realm of end-to-end implicit learning methods for predicting vehicle trajectories.
    \item Implementation of \textbf{graph-based methods with positional encoding for enhanced spatial feature representation}, capturing complex relationships in the scene, which enables LSTM models to effectively capture temporal dynamics.
\end{itemize}

%%%%%%%%%%%%%%%%%%%%%%%%%%%%%%%%%%
% Graphics and Equations
%%%%%%%%%%%%%%%%%%%%%%%%%%%%%%%%%%
%\section{Research background}
\section{RESEARCH BACKGROUND}
\vspace*{1mm}
Numerous research investigations have explored the important role of trajectory prediction in enhancing the safety of autonomous vehicles \cite{bib1}. The author \cite{bib2} explains how to employ Recurrent Neural Networks (RNNs) to predict the future paths of nearby objects in complex driving situations.
%Notably \cite{bib2} embarked on a groundbreaking exploration employing Recurrent Neural Networks (RNNs) to meticulously anticipate the future paths of surrounding entities within intricate driving environments. 
Their model, fine-tuned with real-world driving datasets, yielded promising trajectory prediction outcomes%, thus elevating the safety and situational awareness of autonomous vehicles. 
Another relevant research \cite{bib3} explored using Generative Adversarial Networks (GANs) for probabilistic trajectory prediction. By utilizing GANs, researchers were able to produce various likely future trajectories for vehicles. 

The author \cite{lefevre2014survey} classified vehicle behavior prediction models into three categories: physics-based, maneuver-based, and interaction-aware models. Physics-based models, characterized by the lowest degree of complexity, are constrained to short-term and unreliable predictions. Additionally, these models entirely overlook drivers' tendencies, leading to a decrease in their reliability. Maneuver-based models address this issue by considering each driver's actions as a separate maneuver, independent of other traffic elements. Typically, predictions based on maneuvers are more dependable over an extended period. However, when the number of dimensions in the feature space is increased, it becomes notably more challenging to categorize and predict outcomes accurately.

Furthermore, the work of \cite{bib4} 
%underscored the importance of considering 
considered the interaction dynamics between vehicles and pedestrians. These researchers introduced a novel interaction-aware trajectory prediction model \cite{bib5} proficient in capturing the interplay and dependencies among diverse road users. Through the integration of interaction information, this model demonstrated enhanced precision \cite{bib6} and reliability in trajectory prediction.%, thus contributing significantly to safety in autonomous driving scenarios.

In our research, we aim to tackle the limitations encountered in prior studies. Our focus centers on incorporating a critical element - pedestrians at crosswalks - within a minimally sized yet balanced dataset that encompasses all critical scenarios. To address these challenges, we have carefully crafted a new dataset with annotation information using the Carla simulation platform. By leveraging this synthetic dataset, we anticipate that our model will demonstrate promising performance. This innovative approach empowers us to overcome constraints associated with data collection and processing, ultimately enhancing the outcomes of our study.

%Ego vehicle trajectory prediction is a critical aspect of autonomous driving systems, and several approaches have been explored in prior research to address this complex challenge. The following is a summary of key prior art in this field:
%\begin{itemize}

%\item \textbf{Recurrent Neural Networks (RNNs)}: Early efforts in ego vehicle trajectory prediction often relied on Recurrent Neural Networks (RNNs). These networks can capture temporal dependencies in sequential data, making them suitable for modeling vehicle trajectories. However, RNNs have limitations in handling long-range dependencies and complex spatial relationships.
%\item \textbf{Convolutional Neural Networks (CNNs)}: CNNs have been employed to process sensor data, such as camera images and LiDAR point clouds. They excel in feature extraction from spatial data but may not capture temporal dynamics effectively.
%\item \textbf{Recurrent Convolutional Neural Networks (RCNNs)}: RCNNs combine the strengths of both RNNs and CNNs by using CNNs for feature extraction and RNNs for temporal modeling. These models have shown promise in improving trajectory prediction accuracy.
%\item \textbf{Generative Adversarial Networks (GANs)}: GANs have been explored for trajectory prediction by generating future trajectories conditioned on past observations. Conditional GANs can produce diverse and realistic trajectory samples, enhancing prediction reliability.
%\end{itemize}

% \section{Methodology}
\section{METHODOLOGY}
\vspace*{1mm}
In this section, we explore our model's operation, starting with dataset creation. We detail the use of the Segmentation Anything model (SAM) to generate object candidates. Object-specific features extracted by Deep Neural Networks (DNNs) are then enhanced with a Graph Neural Network (GNN), creating an efficient embedding capturing both object features and spatial relationships. This embedding is employed by a Long Short-Term Memory (LSTM) layer for understanding spatial-temporal features, significantly improving trajectory prediction accuracy.

\subsection{Proposed Approach}
\vspace*{1mm}
We derived a Bird's Eye View (BEV) from the Carla simulation, which offers the flexibility to design various scenarios and control the number of objects within a scene.  Our research presents a comprehensive approach that harnesses the power of computer vision and deep learning to enhance ego vehicle trajectory prediction. Our approach employs a semantic segmentation model for BEV images, providing bounding box coordinates and segmentation masks. Using a DNN network for feature extraction, we create a graph representation of the scene, where nodes are bounding box features connected by edges determined through a K-nearest neighbors algorithm, focusing on spatial relationships. A GNN is utilized to analyze these spatial features. Subsequently, LSTM networks interpret temporal features, enabling precise trajectory forecasting for autonomous vehicles, thus enhancing safety and efficiency, as shown in Figure \ref{fig:image2}.

\subsection{Dataset}
\begin{figure}[h!]
\centering
\includegraphics[width=\columnwidth]{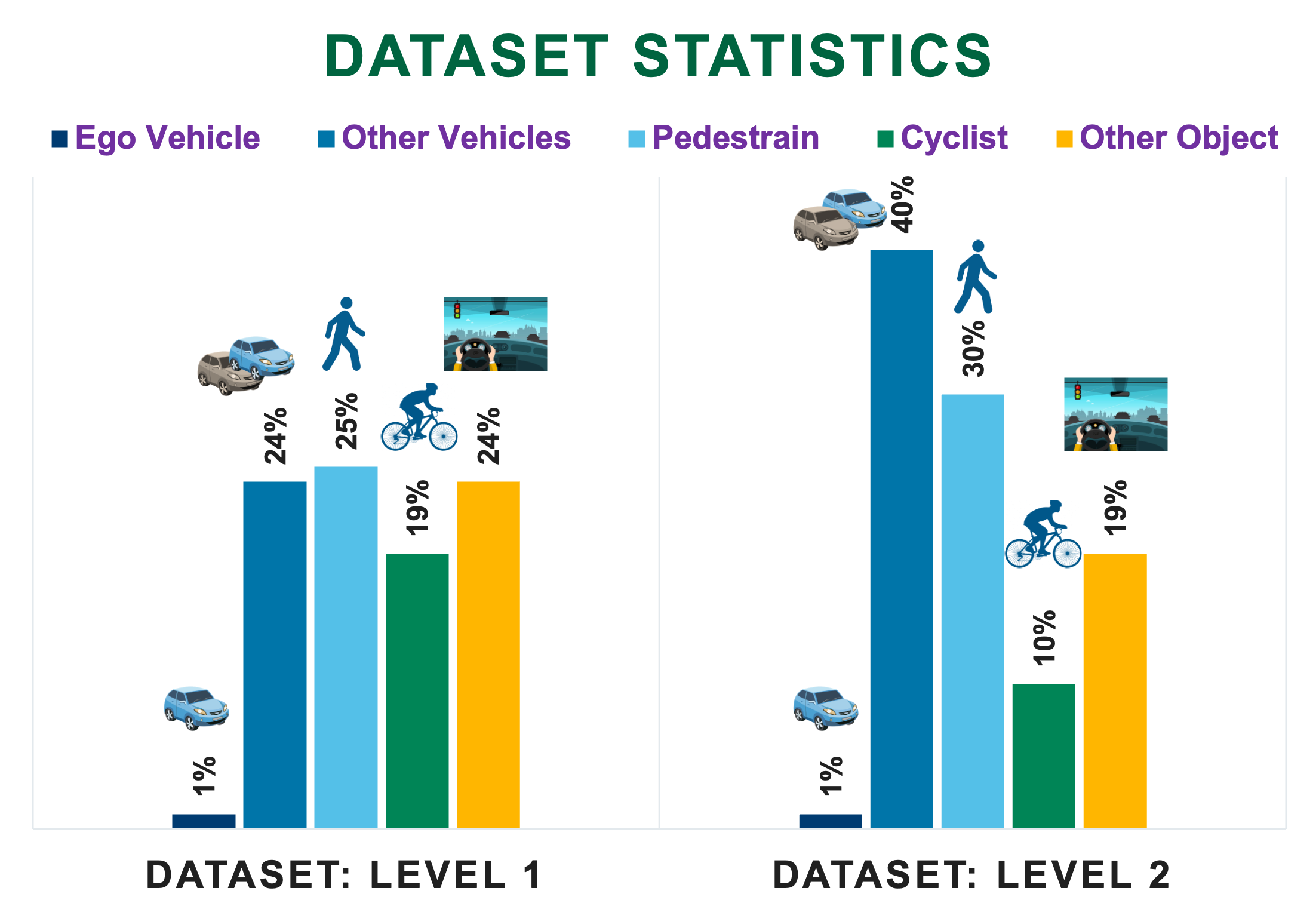}
\vspace*{1mm}
\caption{\textbf{The dataset encompasses two levels}: Level 1, with 1000 images, provides a visual representation of object distribution within the scene, organized by class, while Level 2, consisting of 5000 images, further showcases the representation of objects within the scene categorized by class.}
\label{fig:boschVirtualVisor}
\end{figure}

\vspace*{1mm}
%In this section, we outline the procedures involved in creating datasets\footnote{\textcolor{darkgray}{\href{https://drive.google.com/drive/folders/1JPb64bGV88ymZkJrUBaKQg12tToZVF7T?usp=sharing}{BEV Dataset Link}}} using the CARLA simulator. %To capture comprehensive 360\textdegree  and Bird's Eye View (BEV) perspectives of each scene, the camera position in the simulator was adjusted to achieve a top-down view. This approach, utilizing top-view images for trajectory prediction in autonomous vehicles, facilitates a thorough understanding of the surrounding environment, aiding in precise decision-making.\\ 
In this work, we use the same dataset as in \cite{sharma}\footnote{\textcolor{darkgray}{\href{https://drive.google.com/drive/folders/1JPb64bGV88ymZkJrUBaKQg12tToZVF7T?usp=sharing}{BEV Dataset Link}}}. 
Each image in the dataset has dimensions of 800 pixels in width and 600 pixels in height, with a camera field of view (FOV) set to 90\textdegree. %The CARLA camera is positioned with orientation cam\_rotation = (-90\textdegree, 0\textdegree, -90\textdegree) and location cam\_location = $\mathsf{(0,0,15)}$. 
This configuration, placing the camera 15 meters above the host vehicle, generates a bird's eye view perspective of the vehicle's surrounding.
The dataset has two proposed levels, ``Level 1'' (1000 images) and ``Level 2'' (5000 images), with ``Level 2'' consisting of some more challenging scenes.
%The initial dataset, denoted as ``Level 1," comprises 1000 perspective view images. To enhance realism, ego vehicles and pedestrians were integrated into each scene, and each image was annotated with additional details, including speed and local coordinates (x, y, and z).\\
%Similarly, the subsequent and more challenging dataset, referred to as ``Level 2," consists of 5000 images. To introduce complexity and diversity, the number of vehicles and pedestrians in each scene was increased. This deliberate augmentation was aimed at ensuring the machine-learning models could effectively handle a wide range of real-life scenarios.
All images in this dataset were annotated with class information, with the distributions depicted in Figure \ref{fig:boschVirtualVisor}.
%A novel dataset\footnote{\href{https://drive.google.com/drive/folders/1JPb64bGV88ymZkJrUBaKQg12tToZVF7T?usp=sharing}{Dataset: https://drive.google.com/drive/folders/1JPb64bGV88ymZkJrUBaKQg12tToZVF7T?usp=sharing}}to encourage the research community to pursue the direction of end-to-end implicit vehicle trajectory prediction learning methods.

\subsection{Spatial feature extraction}
%\vspace*{1mm}
In our study, we applied the Segment Anything Model (SAM) \cite{sam} to the dataset we generated to obtain class-specific masks. The bounding box data was utilized to distinguish different objects within the scene. To isolate our ego vehicle, we converted coordinates from the Carla simulation into image coordinates using the camera's extrinsic parameters. This process enabled us to accurately identify class-specific masks along with their corresponding coordinates. For the extraction of object features, we employed three distinct types of DNNs: ResNet18 \cite{resnet}, DenseNet121 \cite{densenet}, ViT \cite{vit}, and EfficientNetB0 \cite{efficientnet}. The primary goal of this approach is to assess and compare the impact of feature sizes derived from these different architectures on the performance of our downstream tasks.

\subsection{Graph Creation}
\vspace*{1mm}
In the process of graph construction, our main aim is to augment the spatial features present in the scene and encode their relational attributes, which are then learned during training. To achieve this, we utilize the features of the objects and their positional coordinates within the image frame, incorporating this information into the graph structure. In our approach, we utilize a K-Nearest Neighbors (KNN) algorithm \cite{knn} to ascertain neighborhood information for each object within the scene. This enables us to establish connections between objects through edges. Importantly, for the weights of these edges, we employ the inverse of the Euclidean distance between objects \cite{gcn_euc}. Mathematically, if the distance between two objects \( i \) and \( j \) is denoted by \( d_{ij} \), the weight of the edge connecting them, \( w_{ij} \), is given by:
\begin{equation}
    w_{ij} = \frac{1}{d_{ij}}
\end{equation}
This method has a key strength: by inversely correlating edge weight with distance, it ensures that closer objects have a stronger connection in the graph. This is reflective of many real-world scenarios where proximity often implies greater interaction or similarity, making this approach particularly effective for accurately modeling spatial relationships and interactions within the scene.  

\subsection{Positional Encoding of Nodes in Graph}
\vspace*{1mm}
In our graph, we have encoded object features and node features along with their distances. However, to enable the model to learn the underlying patterns of spatial relationships between various objects present in a scene, we augment the nodes with positional encoding \cite{posenc}. This approach helps to ensure that similar features are encoded closely together.

Given a dimension \texttt{dim} and a maximum length \texttt{max\_len}, the positional encoding matrix \texttt{PE} is computed as follows:
\begin{equation}
\begin{aligned}
& \texttt{PE}_{(\text{pos}, 2i)} = \sin\left(\text{pos} \cdot 10^{-\frac{2i}{\texttt{dim} \cdot \log(10000)}}\right) \\  
& \texttt{PE}_{(\text{pos}, 2i+1)} = \cos\left(\text{pos} \cdot 10^{-\frac{2i}{\texttt{dim} \cdot \log(10000)}}\right)
\end{aligned}
\end{equation}
where:
\begin{itemize}
    \item \texttt{pos} is a vector with values from 0 to \texttt{max\_len} - 1, representing the position in the sequence.
    \item \( i \) ranges from 0 to \( \frac{\texttt{dim}}{2} - 1 \).
    \item \texttt{PE} is the positional encoding matrix where each row corresponds to a position, and each column corresponds to a dimension of the encoding.
\end{itemize}
This enhanced node feature is then further used for downstream tasks.

\subsection{Graph neural networks}
\vspace*{1mm}
GNNs have emerged as a powerful tool for learning representations on graph-structured data. They extend traditional neural network concepts to the domain of graphs, enabling the effective handling of relational and structural information \cite{gnn_paper_architecture}. GNNs operate by leveraging the connections and features of nodes within a graph, making them particularly adept at tasks like node classification, link prediction, and graph classification \cite{gnn_paper_architecture}.
One of the fundamental mechanisms in GNNs is the aggregation of neighbor information for each node. The basic update rule for a node \(v\) in a GNN can be expressed as: 
\begin{equation}
    h_v^{(k+1)} = \text{UPDATE}\left( h_v^{(k)}, \text{AGGREGATE}\left( \{ h_u^{(k)} : u \in \mathcal{N}(v) \} \right) \right)
\end{equation}
Here, \(h_v^{(k)}\) represents the feature vector of the node \(v\) at the \(k\)-th iteration, \(\mathcal{N}(v)\) denotes the set of neighbors of \(v\), and AGGREGATE and UPDATE are functions that aggregate neighbor features and update the node's feature, respectively \cite{gcn}. This iterative process allows each node to gather information from its local neighborhood and progressively learn more complex features representing the structure of the graph.
\\
In the realm of trajectory prediction, the utilization of GNNs marks a significant advancement \cite{gnn_traj}. The core objective of employing GNNs in this domain is to generate a comprehensive graph embedding that encapsulates spatial information, which is pivotal for understanding and predicting movement patterns. This is achieved by harnessing the power of object-feature interactions and their proximity to each other within a given scene.
 
\subsection{Learning temporal changes using LSTM}
\vspace*{1mm}
 GNNs have limitations in handling temporal features effectively \cite{gnn_temp_t}. This issue arises because GNNs are primarily designed to process spatial relationships within graph-structured data. They excel at capturing the complex interdependencies and hierarchical structures present in graphs. However, when it comes to temporal dynamics – changes that occur over time – GNNs may not inherently capture these aspects as efficiently \cite{gnn_timeseries}.

Temporal features in data are crucial in many real-world applications like time-series forecasting, dynamic network analysis, and trajectory prediction. These applications require understanding not only the spatial relationships among data points but also how these relationships evolve over time. Traditional GNNs might struggle with this because they typically operate on a static snapshot of data, lacking mechanisms to encode changes or movements over time directly. To overcome this we have used LSTM cells along with GNNs graph embedding. LSTM \cite{lstm} networks are a type of recurrent neural network (RNN) particularly adept at learning from sequences of data. They are designed to overcome the limitations of traditional RNNs, such as difficulty in learning long-range dependencies. An LSTM cell consists of multiple gates that regulate the flow of information, allowing it to effectively retain or forget information across long sequences.

The integration of LSTM with GNN embeddings presents a powerful combination for analyzing data with both spatial and temporal components. GNNs excel at generating embeddings that encapsulate the spatial features of data within a graph structure. These embeddings capture the relationships and interactions among nodes in a graph at a given moment.

On the other hand, LSTM cells can learn the temporal dynamics of sequences. When combined with GNN embeddings, LSTMs can interpret how the spatial relationships in the graph evolve over time. This is particularly useful in scenarios where it is crucial to understand not only the structure of the data but also how that structure changes. The general update rule for an LSTM cell can be expressed as follows:
\begin{equation}
    \begin{aligned}
    f_t &= \sigma(W_f \cdot [h_{t-1}, x_t] + b_f) \\
    i_t &= \sigma(W_i \cdot [h_{t-1}, x_t] + b_i) \\
    \tilde{C}_t &= \tanh(W_C \cdot [h_{t-1}, x_t] + b_C) \\
    C_t &= f_t * C_{t-1} + i_t * \tilde{C}_t \\
    o_t &= \sigma(W_o \cdot [h_{t-1}, x_t] + b_o) \\
    h_t &= o_t * \tanh(C_t)
    \end{aligned}
\end{equation}
where \( \sigma \) represents the sigmoid function, \( f_t \) is the forget gate, \( i_t \) is the input gate, \( \tilde{C}_t \) is the candidate cell state, \( C_t \) is the cell state, \( o_t \) is the output gate, and \( h_t \) is the output vector of the LSTM cell. \( W \) and \( b \) are the weights and biases for each gate, respectively, and \( [h_{t-1}, x_t] \) denotes the concatenation of the previous hidden state and the current input.

By applying LSTM cells to the embeddings generated by GNNs, we can effectively learn the changes in the sequences formed from different snapshots of the graph, enabling a deeper understanding of the temporal dynamics in the data that we use for trajectory prediction.

\begin{figure*}[ht]
    \centering
    \includegraphics[width=0.96\textwidth]{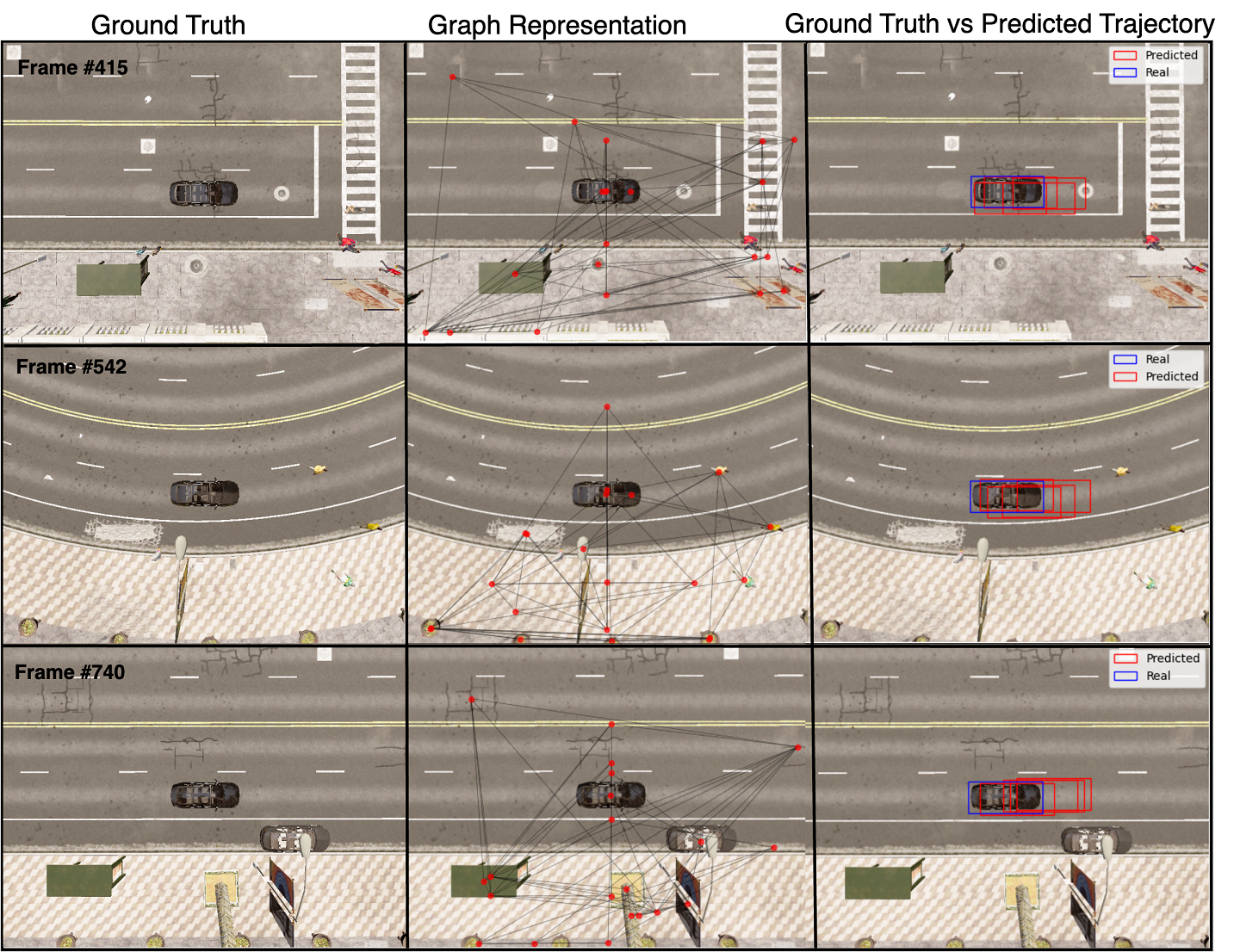}
    \vspace*{2mm}
    \caption{\textbf{Qualitative results:} Left depicts a random frame from the Carla simulation, and the middle illustrates the GNN graph structure with nodes and edges. On the right, the predicted trajectory of the ego vehicle is plotted over a 5-step horizon, where the starting step is marked in blue, and the predicted trajectory is represented in red.}
    \label{fig:image4}
\end{figure*}

\vspace*{5mm}

\begin{table*}[h!]
\centering
\renewcommand{\arraystretch}{1}
\begin{tabular}{ll|lr|crc}
\hline
&&\multicolumn{1}{c}{\textbf{DNN-LSTM}}
&&\multicolumn{2}{c}{\textbf{GNN-LSTM}}\\
\cline{3-6}
\textbf{Backbone} & \textbf{Feature Size} & \textbf{\#Parameters} & \textbf{MSE} & \textbf{\#Parameters} & \textbf{MSE}\\\hline
DenseNet121 \cite{densenet} & 1024 & 7544962 & 3.85 & 174466 & 0.2079\\ 
ResNet18 \cite{resnet} & 512 & 11505474 & 5.82 & 141698 & 0.3385\\
EfficientNet80 \cite{efficientnet} & 1280 & 4729726 & 3.96 & 190850 & 0.1062\\
ViT \cite{vit} & 768 & 86443776 & \textbf{1.87} & 158082 & \textbf{0.0882}\\
\hline
\end{tabular}
\vspace*{3mm}
\caption{\textbf{Comparative Analysis of Backbone Models:} Presented in this table is a detailed comparison of diverse backbone models employed in our study. The analysis includes the integration of DNN with LSTM networks, as well as GNN combined with LSTM. For each model, we have listed both the MSE and the total number of parameters.}
\label{tab:table1}
% \vspace{-0.3cm}
\end{table*}

\section{IMPLEMENTATION DETAILS}
\vspace*{1mm}
%\subsection{Implementation Details}
In our study, we utilized the PyTorch framework for developing our model. For the implementation of the GNN model, PyTorch Geometric \cite{pyg} was employed, with all demonstrated results being derived from the usage of GCN \cite{gcn} layers. Our data was divided into training, testing, and validation sets, with proportions of 80\%, 10\%, and 10\% respectively. To determine the optimal number of layers, learning rate, and weight decay, we engaged Optuna \cite{optuna} for fine-tuning our models. The training process spanned up to 100 epochs and incorporated an early stopping mechanism, set with a tolerance of 10 epochs, to prevent overfitting and ensure efficient training. 

%\section{Experimental Results}
\section{EXPERIMENTAL RESULTS}
\vspace*{1mm}
In this section, we present the performance results of trajectory prediction using both the DNN-LSTM model and our proposed DNN-GNN-LSTM model, as illustrated in Table \ref{tab:table1}. We have documented the MSE errors along with the model sizes to illustrate the effectiveness of our approach.

\subsection{Validation and Evaluation Metrics}
\vspace*{1mm}
The evaluation metrics served as a basis for comparing our approach with other methods. The mean square error (MSE) results, expressed in meters, were reported for each time step (t) within the 5-second prediction horizon. The calculation of MSE at time t follows a standard procedure.
\begin{equation}
\text{MSE} = \frac{1}{n} \sum_{i=1}^{n} (Y_i - \hat{Y}_i)^2    
\end{equation}
where $n$ is the number of elements, $Y_i$ is the real value, and $\hat{Y}_{i}$ is the predicted value.

% Please add the following required packages to your document preamble:
% \usepackage{multirow}

A specific scenario is illustrated in Figure \ref{fig:image4}. For visual representation, a blue bounding box indicates the object's actual position based on the ground truth, while a red bounding box highlights the prediction from our proposed model. Importantly, our model demonstrates outstanding performance in predicting the frame at a 5-step horizon, closely aligning with the corresponding frame at time $t$. Remarkably, it excels even in challenging scenarios, such as navigating a bend where a pedestrian unexpectedly appears. The model effectively captures and comprehends the unpredictable behavior of pedestrians crossing the road, showcasing its exceptional trajectory prediction capabilities.

%\SKS{`Explain figure?}
%\section{Conclusion}

\section{CONCLUSION}
% In this study, we have demonstrated an innovative approach to integrating two distinct model architectures, each with its strengths and limitations, to achieve competitive results in trajectory prediction. A notable aspect of our approach is the efficient fusion of these models, which not only yields optimal trajectory predictions but also does so with a minimal number of parameters. This efficiency makes our model particularly suitable for deployment in resource-constrained devices and environments with computational limitations. Looking ahead, we aim to further refine our trajectory prediction model by incorporating temporal changes into the GNN node features through dictionary learning \cite{tempGCN}. This strategy aligns with a growing trend in the field where researchers are actively exploring ways to enhance the capabilities of GNNs, particularly in addressing their challenges with temporal feature representation. In future, identified a limitation in our current model, specifically the use of top-view images that lack realism. To overcome this challenge, we intend to implement an encoder-decoder transformer. This approach will help us generate a Bird's Eye View (BEV) and improve our trajectory prediction capabilities.
In this study, we have refined our trajectory prediction approach by deconstructing and reevaluating the traditional CNN-LSTM methodology. Previously, we employed a CNN-LSTM model which, while effective, was characterized by a high parameter count, leading to inefficiencies. Recognizing this, we dissected the process into several independent, more manageable stages. This dissection involves initially transforming the image object masks, followed by the utilization of a DNN for feature extraction from the cropped images. The key features are then analyzed using KNN to assess feature relationships and proximity. Subsequently, we construct a graph, embedding positional information for enhanced accuracy. The graph is then processed through a GNN for an optimal graph embedding. Lastly, an LSTM is employed for temporal learning and accurate trajectory prediction. This step-by-step approach not only simplifies the modeling process but also significantly reduces the number of parameters, leading to a more efficient and scalable model. Our future efforts will focus on improving realism in top-view images by implementing an encoder-decoder transformer, aiming to generate a more accurate Bird's Eye View (BEV) and thereby further refine our trajectory prediction model.

%\section{FUTURE WORK}
%\vspace*{1mm}
%Our future efforts will focus on improving realism in top-view images by implementing an encoder-decoder transformer, aiming to generate a more accurate Bird's Eye View (BEV) and thereby further refine our trajectory prediction model.

\section{ACKNOWLEDGMENTS}
\vspace*{1mm}
This publication has emanated from research supported in part by a grant from Science Foundation Ireland under Grant number 18/CRT/6049. For the purpose of Open Access, the author has applied a CC BY public copyright license to any Author Accepted Manuscript version arising from this submission.

%%%%%%%%%%%%%%%%%%%%%%%%%%%%%%%%%%
% Submitting Your Paper
%%%%%%%%%%%%%%%%%%%%%%%%%%%%%%%%%%

%%%%%%%%%%%%%%%%%%%%%%%%%%%%%%%%%%
% Reference Preparation
%%%%%%%%%%%%%%%%%%%%%%%%%%%%%%%%%%

%\section{Acknowledgments} 
%add the acknowledgement section here

% To start a new column (but not a new page) and help balance the last-page
% column length use \vfill\pagebreak.

%%%%%%%%%%%%%%%%%%%%%%%%%%%%%%%%%%
% Bibliography
%%%%%%%%%%%%%%%%%%%%%%%%%%%%%%%%%%

%%%%%%%%%%%%%%%%%%%%%%%%%%%%%%%%%%
% Biography
%%%%%%%%%%%%%%%%%%%%%%%%%%%%%%%%%%

%\begin{biography}
%Please submit a brief biographical sketch of no more than 75 words. 
%Include relevant professional and educational information as shown 
%in the example below.

%Jane Doe received her BS in physics from the University of Nevada (1977) 
%and her PhD in applied physics from Columbia University (1983). Since 
%then she has worked in the Research and Technology Division at Xerox 
%in Webster, NY. Her work has focused on the development of toner adhesion 
%and transport issues. She is on the Board of  IS\&T and a member of APS 
%and SPIE.
% \end{biography}

\end{document}